%% file: dual_arxiv.tex
\pdfoutput=1
\documentclass{llncs}

\usepackage{amsfonts,amsmath}
\usepackage{epigraph}

\usepackage{hyperref}
\usepackage{breakurl}

\input pheader24.tex

\begin{document}
\title{
% Mirrors: an LLM-friendly Logic Reasoner
%Through the Looking Glass, and what Negation Found There
Through the Looking Glass, and what Horn Clause Programs Found There
%Through the Looking Glass, and what Dual Horn Formulas Found There
% 
}
\author{{\bf Paul Tarau}\orcidID{0000-0001-7192-9421}}

\institute{
%   {Department of Computer Science and Engineering}\\
   {University of North Texas}\\
   {\em paul.tarau@unt.edu}
}

\maketitle
%\linenumbers

\epigraph{
``Well, now that we have seen each other,'' said the unicorn, ``if you'll believe in me, I'll believe in you.''
}
{Lewis Carroll, Through the Looking-Glass and What Alice Found There}

\begin{abstract}
Dual Horn clauses mirror key properties of Horn clauses. This paper explores the ``other side of the looking glass'' to reveal some expected and unexpected symmetries and their practical uses.

We revisit Dual Horn clauses as enablers of a form of constructive negation that supports goal-driven forward reasoning and is valid both intuitionistically and classically. In particular, we explore the ability to  falsify a counterfactual hypothesis in the context of a background theory expressed as a Dual Horn clause program.

With  Dual Horn clause programs, by contrast to negation as failure,  the variable bindings in their computed answers provide explanations for the reasons why a statement is successfully falsified.
Moreover, in the propositional case, by contrast to negation as failure as implemented with stable models semantics in ASP systems, and similarly to Horn clause programs, Dual Horn clause programs have polynomial complexity. 

After specifying their execution model with a metainterpreter, we devise a compilation scheme from Dual Horn clause programs to Horn clause programs, ensuring their execution with no performance penalty and we design the embedded SymLP language to support combined Horn clause and Dual Horn clause programs.

As a (motivating) application, we cast LLM reasoning chains into propositional Horn and Dual Horn clauses that work together to constructively prove and disprove goals and enhance Generative AI with explainability of reasoning chains.

\end{abstract}

\section{Introduction}

The concept of negation in logic programming traditionally relies on  ``negation as failure'', which infers the negation of a proposition if the proposition itself cannot be proven. This approach, while useful, often lacks the ability to provide clear explanations for why a proposition is considered false (given the absence of variable bindings). This is a problem, especially in complex scenarios where understanding the reasoning process is crucial. 

In fact, negation as failure tacitly reflects a meta-level property about the proof procedure into the object-language, and as this can be iterated on, the resulting semantics is unavoidably subtle requiring carefully designed guardrails for both practical programming uses and theoretical foundations. Among them, while elegant and expressive, the stable-models  semantics \cite{GelfondL88} based handling of negation in Answer Set Programming (ASP) is already NP-complete even in the propositional case \cite{dantsin} and must rely on grounding with exponential space complexity in the case of Datalog programs.

Curiosity about the possibility of a simpler and more informative negation mechanism in Logic Programming brings us to our ``back to the future'' revisiting of Dual Horn clauses.

Dual Horn clauses have emerged as an interesting set of propositional formulas
in Schaefer's ``{\em dichotomy theorem}'', that classifies propositional formulas in P-complete vs. NP-complete classes \cite{schaefer78}, by
falling in the P-complete class together with their Horn clause cousins.

Surprisingly, while their theoretical properties have been well known for very long time, we have found no evidence of their uses in programming or knowledge representation tasks and no mentions of uses in more expressive logic languages beyond the propositional case.

In part, this motivates our curiosity for revisiting them simply as logic programming expressiveness enhancers. Another motivation comes from the ability to generate interesting propositional Dual Horn clause programs with LLMs, 
%along the lines of \cite{flops24,tarau2023automation}, 
when the problem we focus on involves forward reasoning from a fact to its consequences. This contrasts to Prolog's usual backward reasoning elaboration of a goal into progressively more solvable alternatives and subgoals.

As we will show next, it turns out that the significance of Dual Horn clauses centers around their ability to constructively handle negation, moving beyond the limitations of negation as failure. 
By enabling {\em goal-driven forward reasoning}, Dual Horn clauses support the exploration and falsification of hypotheses in the presence of negative factual information. This capability is particularly valuable in fields where reasoning about counterfactuals and understanding the underlying logic of decisions are important.
With constructive logic in mind, we will also point out the properties shared between Horn clause and Dual Horn clause programs that also hold in Intuitionistic Logic when represented in implicational form ({\bf section} \ref{bgr}).

After describing a metainterpreter, we devise a compilation scheme that transforms Dual Horn clause programs into conventional Horn clause programs allowing their execution with no performance penalty. 
This development ensures that the enhanced capabilities of Dual Horn clauses can be seamlessly integrated into existing logic programming systems, leading to a sketch of an embedded language accommodating Horn clauses and Dual Horn clause working together, {\tt SymLP} ({\bf section} \ref{sym}). 
We will  describe, using the notations of {\tt SymLP} several examples of reasoning patterns including default reasoning, theory falsification, decision on conflicting information and interactions with negation as failure ({\bf section} \ref{pats}).

We will also expand the utility of Dual Horn clauses by exploring their application in the context of Generative AI and Large Language Models (LLMs). By casting the reasoning steps of LLMs into propositional Horn and Dual Horn clauses, we provide a framework that not only supports the proving or disproving of goals but also enhances the explainability of the LLMs' decision-making processes. The ability to explain AI decisions transparently is crucial for building trust and for the practical deployment of AI systems in sensitive or impactful domains ({\bf section} \ref{llms}).

After discussing related work ({\bf section} \ref{rel}) we will conclude the paper and discuss future work ({\bf section} \ref{conc}).

The SWI-Prolog code described in the paper is available online\footnote{
\url{https://github.com/ptarau/TypesAndProofs/tree/master/symlp}}.

\begin{comment}
\BI

\I both Horn and Dual Horn are polynomial in the propositional case (vs. NF as handled by NP complete stable models)
\I both Horn and Dual Horn are simply exponential in the Datalog case (after grounding)
vs. double exponential as for NP complete stable models)
\I intuitionistic and classical semantics agree on the proposed subset
\I relation  to constructor theory: separate things that are possible and things that are impossible - each with their own logic
\I relation to LLMs: ref. to the recursors paper: task decomposition vs. (bad) consequence prediction
\I LLMs can generate large number of positive and negative facts
\I interest in LLM generated propositional logic is also justified by the direct use of vector-embeddings based soft unification
\EI
\end{comment}

\section{Background on Dual Horn clauses}\label{bgr}

A {\em Horn clause} is a disjunction of literals with at most one positive literal.
A {\em Definite Horn clause} is a disjunction of literals with exactly one positive literal.
A {\em Dual Horn clause} is a disjunction of literals with at most one negative literal.
A {\em Definite Dual Horn clause} is a disjunction of literals with exactly one negative literal.
{\em A Horn clause Program} %(or Horn clause Formula) 
is a conjunction of Horn clauses.
A {\em Dual Horn clause Program}  is a conjunction of Dual Horn clauses.

The following formulas describe Horn clauses (1) and (2), vs. Dual Horn clauses (3) and (4). Note the disjunctive forms as expressed, for instance, in resolution theory and the implicational forms as expressed in Prolog or Answer Set Programming (ASP) programs.
\begin{center}
{\em Horn}
\BEQ
p_0 \OR \NOT  p_1 \OR \ldots \OR \NOT  p_n.
\EEQ
\BEQ
p_0 \IF p_1 \AND \ldots \AND p_n.
\EEQ
~\\
{\em Dual Horn}
\BEQ
\NOT p_0 \OR  p_1 \OR \ldots \OR   p_n.
\EEQ
\BEQ
p_0 \FI p_1 \OR  \ldots \OR p_n.
\EEQ
\end{center}
Similarly, the two forms make sense when extending Definite Horn Horn clauses  with integrity constraints called {\em denials}  \cite{eshghi89} (see (5) and (6) ), asserting that at least one $p_i$ must be false
and their duals (see (7) and (8)), asserting that at least one $p_i$ must be true.

%, but still polynomial: Integrity constraints
\begin{center}
 Horn
\BEQ
\NOT  p_1 \OR \ldots \OR \NOT  p_n.
\EEQ
\BEQ
false \IF p_1 \AND \ldots \AND p_n.
\EEQ
~\\
Dual Horn
\BEQ
p_1 \OR \ldots \OR  p_n.
\EEQ
\BEQ
true \FI p_1 \OR  \ldots \OR p_n.
\EEQ
\end{center}

While the usual assumption about the underlying logic in Prolog and SAT solvers is classical logic (CL) or a fairly close intermediate logic in the case of ASP systems, the proof-theoretical semantics \cite{miller21} of Horn clause Logic in implicational form has been known to be compatible also with its reading in Intuitionistic Logic (IL). We will keep this in mind when covering some key properties of both Horn clause and Dual Horn clause programs. %The following proposition holds:
\BP
\BI
\I (1) implies (2) in IL and (1) is equivalent to (2) in CL.
\I (3) implies (4) in IL and (3) is equivalent to (4) in CL.
\I (5) implies (6) in IL and (5) is equivalent to (6) in CL.
\I (7) and (8) are equivalent both in IL and CL.
\EI
Proofs of the CL statements are trivial using De Morgan and definition of material implication.
The implicational forms in IL are weaker given that De Morgan  holds  only one way. Proofs in IL are otherwise easy using a sequent calculus system like \cite{dy1} or its corresponding sound and complete theorem prover \cite{tplp22}.
\EP

\BP
Satisfiability of Horn clause and Dual Horn clause formulas with or without integrity constraints is P-complete.
\EP

This follows from Schaefer's dichotomy theorem classifying propositional formulas \cite{schaefer78} in P-complete vs. NP-complete types. 
Note  that finding a renaming that might turn a set of clauses into a set of Horn or Dual Horn clauses are also polynomial \cite{renhorn90}.

\BX We start with a small Dual Definite program, adopting
Prolog-like syntax, with $\rightarrow$ represented as ``\verb~=>~'' and $\OR$ represented as ``;''.

\begin{codex}
p => q ; r.
q => r ; s.
r => false.
s => false.
\end{codex}
\EX
Note also that ``\verb~s => false~'' represents a negated fact (in either CL or IL), the same way as ``\verb~s :- true~'' would represent a positive fact.
Let's proceed with {\tt p} as our goal, similarly as if we would evaluate a Horn clause program. Assuming that {\tt p} were  true, we would infer that at least one of {\tt q} , {\tt r} and {\tt s} should be true.That reduces to the truth of {\tt r} and {\tt s} and finally the truth  of {\tt s}. But {\tt s} implies {\tt false},  which then backpropagates, falsifies {\tt s} and {\tt r} and then falsifies the initial goal {\tt p}. Note that this reasoning is also intuitionistically valid similarly to its Horn clause counterpart.

We conclude from this example that we have a goal oriented {\em falsification} process for Dual Horn programs that mimics {\em verification} in Horn Programs via SLD-resolution. It is easy to see that programs with variable bindings generated via unification will work also in a similar way. 

More generally, the falsification process relies on the following fact:
\BP The relations (9)  and (10) hold both in IL and CL.
\begin{center}
\BEQ
(p_0 \IF p_1 \AND \ldots \AND p_n)  \AND p_1 \AND \ldots \AND p_n  \FI p_0
\EEQ

\BEQ
(p_0 \FI p_1 \OR \ldots \OR p_n)  \AND \NOT p_1 \AND \ldots \NOT p_n  \FI  \NOT p_0
\EEQ
\end{center}
\EP

\noindent Thus, based on (10),  to falsify $p_0$ we need to falsify all the disjuncts $p_i$ that are consequences of $p_0$, the essence of a goal-driven falsification process, operationally similar to SLD-resolution proof procedure on Horn clause programs as illustrated by (9).

\section{Symmetric Logic Programming (SymLP) : the two sides of the mirror, together}\label{sym}

The next step is to design a mechanism for the ``safe  cohabitation'' of Horn clauses and Dual Horn clauses in the same program.
To ensure that their predicates defining rules and facts are disjoint (for instance, to avoid contradictions) we will place
the Horn clause component in module {\tt true} and the Dual Horn clause component in module {\tt false}.
Note that using modules, is mostly a syntactic simplification for implementing them in Prolog, as any mechanism ensuring that the predicate symbols are distinct would do (e.g., by prefixing predicate names with symbols {\tt true} and {\tt false}).

\subsection{Syntax} To embed the SymLP language that combines Horn clauses and Dual Horn clauses into Prolog, we will use a few operators. We use ``{\tt +}'' to mark  facts that are true  and ``{\tt -}'' to mark facts that are false. We will use ``\verb~=>~'' to represent implication and ``\verb~<=~'' to represent reverse implication. 
Note also that \verb~-p~ can be seen as a shortcut for \verb~p=>false~ and that similarly, {\tt +p}  can be seen as a shortcut for \verb~p<=true~. We will borrow from Prolog the usual notation for conjunction ``{\tt,}'' and disjunction ``{\tt ;}''.

\subsection{Implementing Symmetric Logic Programs}

Next, we will describe a surprisingly simple mechanism to implement a goal-oriented execution mechanism for Dual Horn programs, inspired by the SLD-resolution mechanism of Prolog.

\subsection{The dual use meta-interpreter}

We start with a simple difference-list based metainterpreter that will turn out to work for both Horn clause and Dual Horn clause programs.

\begin{code}
metaint([]).        % no more goals left, succeed
metaint([G|Gs]):-   % unify the first goal with the head of a clause
   cls([G|Bs],Gs),  % build a new list of goals from the body of the  
                    % clause extended with the remaining goals as tail
   metaint(Bs).     % interpret the extended body 
\end{code}

The following examples show that the representation provided as the predicate {\tt cls/2} works in both cases.

\BX Simple Horn program with Prolog-like notation:
\begin{code}
p <= q,r.
q <= r,s.
r <= true,
s <= true.
\end{code}

Difference list representation of clauses:
\begin{codex}
cls([ p,q,r     |Tail],Tail).
cls([ q,r,s     |Tail],Tail).
cls([ r         |Tail],Tail).
cls([ s         |Tail],Tail).
\end{codex}

Successful verification:
\begin{codex}
?- metaint([p]).
true
\end{codex}
\EX

\BX Dual Horn program with \verb~=>~ as implication and {\tt ;} as disjunction.
\begin{code}
p => q ; r.
q => r ; s.
r => false.
s => false.
\end{code}
We can use the {\em same} difference list representation of Dual Horn clauses:
\begin{codex}
cls([ p,q,r     |Tail],Tail).
cls([ q,r,s     |Tail],Tail).
cls([ r         |Tail],Tail).
cls([ s         |Tail],Tail).
\end{codex}
Successful falsification is now computed as well as:
\begin{codex}
?- metaint([p]).
true
\end{codex}
\EX

\BP
Falsification of a goal of a Dual Horn clause program by the metainterpreter succeeds if and only if the proof of the goal with a corresponding Horn Clause program succeeds.

{\em Proof}: By flipping each literal in formula (3), formula (1) is obtained, which is a Horn clause. Note that success of the interpreter on (1) corresponds to its success on (3). 
\EP

Note that the meta-intepreter also accommodates terms with variables, in which case
variable bindings will be  informative about {\em why something is successfully falsified}.

% TODO: in the propositional case, proof traces

\subsection{The Compilation to Prolog clauses}\label{compile}

Observing that after translation to a uniform clause representation the same meta-interpreter works for Horn programs and their duals, suggests that a simple compilation scheme from Dual Horn clauses to Horn clauses must exist, and when extended to the case of SymLP programs, it will just keep Horn clauses invariant.

We will implement it here using Prolog's {\tt term\_expansion}\footnote{
\url{https://www.swi-prolog.org/pldoc/man?predicate=term_expansion/2}
} that overloads the Prolog reader with a call to a {\tt compile\_clauses(SymLPclause,EquivalentPrologclause)}.
The key idea is to convert Dual Horn clauses occurring in a Prolog program to corresponding Horn clauses that, when executed, follow the semantics specified
by the dual-use meta-interpreter.

We will distinguish successful falsification from successful proof of the ``compiled'' Horn program simply by placing the result of the {\tt term\_expansion/2} of Horn clauses into the module {\tt true} and the result for Dual Horn clauses into the module {\tt false}.

After defining ``\verb|<=|'' (reverse implication) and ``\verb|=>|'' (implication) as operators,
the predicate {\tt compile\_clauses} will be called at term expansion time to convert the 
Dual Horn clauses to Horn clauses to be placed in module ``{\tt false}``.
Note also that in the bodies of Dual Horn clauses disjunctions will be converted to conjunctions. 
Facts marked with ``\verb~-~'' will  be placed in module {\tt false}.
The action on Horn clauses introduced by `` \verb~<=~'' is  just replacing 
 `` \verb~<=~" with `` \verb~:-~" to be then added together with facts marked with  `` \verb~+~" to the module  ``{\tt true}''.

\begin{code}
:- multifile(term_expansion/2).

:-op(1199,xfx,(=>)).
:-op(1199,xfx,(<=)).
\end{code}

\begin{code}
compile_clauses(C,_):-var(C),!,fail.

compile_clauses((H<=B),true:(H:-B)):-!,nonvar(H),nonvar(B).
compile_clauses((+H),true:H):-!,nonvar(H).

compile_clauses((H=>B),R):-nonvar(H),nonvar(B),!,dual2clause((H=>B),R).
compile_clauses((-H),false:H):-nonvar(H).
\end{code}

\begin{code}
dual2clause((H=>false),false:(H)):-!.
dual2clause((H=>B),false:(H:-CB)):-disj2conj(B,CB).
dual2clause((-H),false:H).

disj2conj((A;B),(CA,CB)):-nonvar(A),nonvar(B),!,
    disj2conj(A,CA),
    disj2conj(B,CB).
disj2conj(A,A).
\end{code}

\BX As the Horn clauses get compiled simply by replacing \verb~<=~ with \verb~:-~, we will illustrate next what happens to a Dual Horn clause program:
\begin{code}
:-include('compile_clauses.pro').

p => q ; r.
q => r ; s.
r => false.
s => false.
\end{code}

\noindent becomes:

\begin{code}
% in module false
p :- q , r.
q :- r , s.
r :- true.
s :- true.
\end{code}
Then, querying it with:
\begin{codex}
?- false:p.
true
\end{codex}
the success confirms that {\tt p} is indeed falsifiable.
\EX

\section{Reasoning patterns expressed in SymLP}\label{pats}
We will next overview a few reasoning patterns expressed as SymLP programs.

\subsection{Default reasoning, a dual view}
\BX Dually to the usual way to express defaults and exceptions, we just declare  which birds are not challenged when defining which of them can fly.
\begin{codex}
:-include('compile_clauses.pro').

fly(X) <= bird(X),false:challanged_bird(X).
\end{codex}

\begin{codex}
+bird(tweety).
+bird(chicken_little).
+bird(eagle_joe).
+bird(humming_jenny).

-challanged_bird(eagle_joe).
-challanged_bird(humming_jenny).
\end{codex}

\begin{codex}
?- true:fly(X).
X = eagle_joe ; X = humming_jenny.
\end{codex}
\EX

\subsection{Dual Horn programs and the logic of theory falsification}

Falsifiability of a theory has been known for a long time \cite{popper34} as instrumental to make the theory predictive and testable, and thus useful in practice.
In terms of Dual Horn programs, this is expressed by saying that something should fail if its consequences fail and by observing that falsity propagates back from false facts to rules that rely to them. This suggest a proof procedure to be applied to Dual Horn clauses  described in formula (4).

\BX We will clarify this by working out an example of ``theory falsification''\footnote{obtained by edits of an LLM request to explain problems with assuming the existence of negative gravity fields}.

\begin{code}
:-include('compile_clauses.pro').

'Negative gravity fields are possible' =>
    'Planets and stars would disperse' ;
    'Atmospheres of planets would be pushed away from their surfaces' ;
    'Unresolvable paradoxes in physics'.

'Planets and stars would disperse' => false.
\end{code}
\begin{code}
'Unresolvable paradoxes in physics' =>
   'Relativity theory is incorrect' ;
   'Quantum field theory is incorrect'.

'Relativity theory is incorrect' => false.
'Quantum field theory is incorrect' => false.

'Atmospheres of planets would be pushed away from their surfaces' => false.
\end{code}
The result of the ``execution'' of this Dual Horn clause program could be expressed by a query goal of the form:
\begin{codex}
?- false:'Negative gravity fields are possible'.
true
\end{codex}
\EX
confirming that ``{\tt Negative gravity fields are possible}'' has been successfully falsified in the context of our accepted background knowledge about physics.

\subsection{Combining positive and negative advice in a SymLP program}

The next example of SymLP program will have calls across the {\tt true} and {\tt false} modules. Note that the {\tt include} statement will trigger the compilation process, as explained in subsection \ref{compile}.

\BX Combining nuances of  (a fictional example) of advice on stocks. \label{stocks}
\begin{code}
:-include('compile_clauses.pro').

cautious_buy(X)<=recommended(X),safe(X).

safe(X) <= false:volatile(X).
safe(X) <= false:overvalued(X).
safe(X) <= true:stable(X).

+recommended(qqq).
+recommended(bitcoin).
+recommended(apple).
+recommended(meta).
+recommended(berkshire).

+stable(att).
+stable(berkshire).

volatile(X) => big_price_changes_last_month(X).

-big_price_changes_last_month(apple).
-big_price_changes_last_month(meta).
-big_price_changes_last_month(comcast).

-overvalued(qqq).
\end{code}

After combining positive and negative facts and inferences drawn from them in Horn module {\tt true} and Dual Horn module {\tt false} the call to {\tt cautious\_buy(X)} in module {\tt true} will generate the following answers:
\begin{codex}
?- true:cautious_buy(X).
X = qqq ;
X = apple ;
X = meta ;
X = berkshire.
\end{codex}
\EX
At this point,  one might want to ask the legitimate question: 
\begin{quote} Why  would we represent negation by explicitly listing negative facts, knowing that something like {\tt not(white(swan,X))} will include not just ``black swans'' but also an infinite set of unrelated entities (e.g., ``globular galaxies'') ? 
\end{quote}
%\noindent as negation as failure under the closed world assumption would happily agree with.

First, like in the stock market advising in Example \ref{stocks}, decisions are justified by a small set of positive or negative facts from where a decision process initiates. 
Next, reasoning with Machine Learning datasets (including those used in Inductive Logic Programming) relies on finite sets of positive and negative facts. %information streams that can be constructively enumerated.
In particular, in Generative AI, a way to control hallucinations is by implementing, as part of a a  multi-agent framework,  generation of positive and negative facts  and rules relying on them, % \cite{flops24}.
as we will show in section \ref{llms}.

\subsection{Negation as failure to prove and affirmation as failure to disprove}

Assuming that we have compiled our SymLP e program into the modules {\tt true} and {\tt false}, negation as failure ({\tt unverifiable/1}) and its dual ({\tt unfalsifiable}) are implemented as follows:

\begin{code}
unverifiable(X):-not(true:X).
unfalsifiable(X):-not(false:X).
\end{code}

Note  that combining modules {\tt true} and {\tt false} into one program has the same complexity as Horn clause programs.
Thus it has a single minimal model that can be computed in polynomial time in the propositional case. 
This holds true also if {\tt not/1} calls between modules {\tt true} and {\tt false} result in a stratified compiled program \cite{eiter2007complexity}.
Otherwise, we are back to the usual pitfalls of Prolog's negation as failure. Therefore, 
a semantically safe and simple use of Prolog's {\tt not/1} would be to only  apply it to predicates in
modules  {\tt true} and {\tt false} from {\em outside} (e.g., from Prolog's module ``{\tt user}'').

\BX Using Prolog's {\tt not/1} in a SymLP program.
\begin{codex}
:-include('compile_clauses.pro').

exonerated(X) <= suspect(X),false:proven_guilty(X).

investigated(X) <= suspect(X),not(false:proven_guilty(X)).

+suspect(alice).
+suspect(bob).

proven_guilty(X) => found_of(X,dna) ; found_of(X,fingerprints).

-found_of(alice,_anything).
\end{codex}
{\em Not falsifiable} that Bob is {\tt proven\_guilty} only entails that he is still investigated (vs. exonerated if not {\em proven} guilty in the case of Alice).
\begin{codex}
?- exonerated(X).
X = alice.

?- investigated(X).
X = bob.
\end{codex}
\EX

\newpage
Let us just mention that, similarly to ASP and s(CASP), concepts  related to epistemic modalities emerge, when combining the independent opinions originating on the two sides:
\BI
\I things that are clearly unknown: failure to prove and failure to disprove
\I things that are strongly accepted as known: successfully proven and failing to disprove
\I things that are very likely to be impossible: successfully falsified and failing to prove.
\EI

\section{LLM generated Dual Horn programs}\label{llms}

Given the ability to steer an LLM to explore recursively a given topic while staying focussed on the objective specified by an initiator goal \cite{flops24,tarau2023automation}, the LLM can be used to generate large sets of high quality positive or negative facts as well as rules describing their inference steps, to be all exported as a Prolog program.

\BX The following is a complete example of a propositional Dual Horn clause 
program\footnote{Ready to be tried out online with the DeepLLM system  at \url{https://deepllm.streamlit.app/}}
 generated from a recursively explored (up to depth = 2)
initiator goal
asking to falsify the misguided belief about: ``{\em escalation risks after use of tactical nuclear weapons}'':
\vskip 0.5cm
\begin{codex}
'escalation risks after use of tactical nuclear weapons'=>
    'Global nuclear war';
    'Uncontrollable retaliation cycles'.
'Global nuclear war'=>
    'Widespread radioactive fallout',
    'Massive civilian casualties',
    'Long-term environmental damage',
    'Global economic collapse',
    'Irreversible climate change',
    'Extensive agricultural failure'.
'Widespread radioactive fallout'=>
    'Long-term environmental damage';
    'Massive civilian casualties'.
\end{codex} 
  
\begin{codex}   
'Long-term environmental damage'=>
    'Persistent ecosystem disruption',
    'Chronic health conditions',
    'Irreversible soil contamination',
    'Permanent loss of biodiversity';
    'Widespread radiation exposure';
    'Persistent ecological disruption';
\end{codex} 
  
\begin{codex}  
'Massive civilian casualties'=>
    'Severe humanitarian crises',
    'Overwhelmed medical systems',
    'Long-term psychological impacts',
    'Economic destabilization';
    'Widespread environmental destruction';
    'Long-term radiation effects'.
'Widespread environmental destruction'=>
    'Long-term habitat loss',
    'Severe biodiversity decline',
    'Persistent soil contamination',
    'Irreversible climate impacts',
    'Chronic water shortages'.
\end{codex} 
  
\begin{codex} 
'Long-term radiation effects'=>
    'Genetic mutations',
    'Chronic health disorders',
    'Environmental contamination',
    'Agricultural degradation'.
'Widespread radiation exposure'=>
    'Genetic mutations increase',
    'Agricultural crop failure',
    'Chronic health deterioration',
    'Wildlife population decline'.
\end{codex} 
  
\begin{codex} 
'Persistent ecological disruption'=>
    'Species extinction rates',
    'Reduced agricultural productivity',
    'Chronic health conditions',
    'Loss of biodiversity'.
\end{codex} 
  
\begin{codex} 
'Global economic collapse'=>
    'Widespread humanitarian crisis';
    'Irreversible environmental damage';
    'Severe job losses',
    'Market instability',
    'Investment crashes',
    'Supply chain disruptions',
    'Increased poverty rates',
    'Currency devaluation'.
\end{codex} 
  
\begin{codex} 
'Widespread humanitarian crisis'=>
    'Massive refugee movements',
    'Severe food shortages',
    'Critical medical deficiencies',
    'Intensified civil unrest'.
\end{codex} 
  
\begin{codex} 
'Irreversible environmental damage'=>
    'Long-term habitat destruction',
    'Permanent loss of biodiversity',
    'Severe air quality deterioration',
    'Unrecoverable soil degradation',
    'Extinction of critical species',
    'Irreversible climate alterations'.
\end{codex} 
  
\begin{codex}  
'Irreversible climate change'=>
    'Widespread ecological collapse';
    'Permanent agricultural devastation'.
'Widespread ecological collapse'=>
    'Massive species extinction',
    'Severe food shortages',
    'Chronic water scarcity',
    'Intensified natural disasters'.
\end{codex} 
  
\begin{codex} 
'Permanent agricultural devastation'=>
    'Widespread famine crisis',
    'Long-term soil infertility',
    'Severe food shortages',
    'Collapse of ecosystems'.
'Extensive agricultural failure'=>
    'Widespread famine crisis';
    'Severe ecological disruption'.
\end{codex} 
  
\begin{codex} 
'Widespread famine crisis'=>
    'Severe malnutrition rates',
    'Increased mortality rates',
    'Economic instability',
    'Social unrest';
'Severe ecological disruption'=>
    'Biodiversity loss',
    'Habitat destruction',
    'Soil degradation',
    'Water resource depletion'.
\end{codex} 
  
\begin{codex} 
'Uncontrollable retaliation cycles'=>
    'Global security destabilization',
    'Widespread humanitarian crises',
    'Escalated military conflicts',
    'Severe economic disruptions'.
'Global security destabilization'=>
    'Unpredictable military responses';
    'Widespread humanitarian crises'.
\end{codex} 
  
\begin{codex} 
'Unpredictable military responses'=>
    'Widespread civilian casualties',
    'Severe economic disruptions',
    'Long-term environmental damage',
    'Sudden geopolitical shifts'.
\end{codex} 
  
\begin{codex} 
'Widespread humanitarian crises'=>
    'Massive displacement waves',
    'Severe resource shortages',
    'Intensified disease outbreaks',
    'Heightened conflict incidents';
    'Global economic collapse';
    'Massive refugee movements';
    'Massive refugee movements',
    'Intensified food shortages',
    'Overwhelmed medical systems',
    'Increased child mortality'.
\end{codex} 
  
\begin{codex} 
'Massive refugee movements'=>
    'Overburdened local resources',
    'Increased social tensions',
    'Strained healthcare systems',
    'Economic destabilization',
    'Environmental degradation'.
\end{codex} 
  
\begin{codex} 
'Escalated military conflicts'=>
    'Global humanitarian crises';
    'Widespread ecological disasters'.
'Global humanitarian crises'=>
    'Widespread famine outbreaks',
    'Mass displacement waves',
    'Severe medical shortages',
    'Intensified poverty levels'.
'Widespread ecological disasters'=>
    'Long-term habitat destruction',
    'Severe biodiversity loss',
    'Persistent soil contamination',
    'Irreversible water pollution'.
'Severe economic disruptions'=>
    'Global market collapse';
    'Widespread humanitarian crises';
'Global market collapse'=>
    'Widespread unemployment surge',
    'Investment capital evaporation',
    'Consumer spending plummet',
    'International trade paralysis',
    'Financial sector instability'.
\end{codex} 
  
\begin{codex} 
-'Persistent ecosystem disruption'.
-'Chronic health conditions'.
-'Irreversible soil contamination'.
-'Permanent loss of biodiversity'.
-'Severe humanitarian crises'.
-'Overwhelmed medical systems'.
-'Long-term psychological impacts'.
-'Economic destabilization'.
-'Long-term habitat loss'.
-'Severe biodiversity decline'.
-'Persistent soil contamination'.
-'Irreversible climate impacts'.
-'Chronic water shortages'.
-'Genetic mutations'.
-'Chronic health disorders'.
-'Environmental contamination'.
-'Agricultural degradation'.
\end{codex} 
  
\begin{codex} 
-'Genetic mutations increase'.
-'Agricultural crop failure'.
-'Chronic health deterioration'.
-'Wildlife population decline'.
-'Species extinction rates'.
-'Reduced agricultural productivity'.
-'Loss of biodiversity'.
-'Severe food shortages'.
-'Critical medical deficiencies'.
-'Intensified civil unrest'.
-'Long-term habitat destruction'.
-'Severe air quality deterioration'.
-'Unrecoverable soil degradation'.
-'Extinction of critical species'.
-'Irreversible climate alterations'.
-'Massive species extinction'.
-'Chronic water scarcity'.
\end{codex} 
  
\begin{codex} 
-'Intensified natural disasters'.
-'Long-term soil infertility'.
-'Collapse of ecosystems'.
-'Severe malnutrition rates'.
-'Increased mortality rates'.
-'Economic instability'.
-'Social unrest'.
-'Biodiversity loss'.
-'Habitat destruction'.
-'Soil degradation'.
-'Water resource depletion'.
-'Widespread civilian casualties'.
\end{codex} 
  
\begin{codex}
-'Sudden geopolitical shifts'.
-'Massive displacement waves'.
-'Severe resource shortages'.
-'Intensified disease outbreaks'.
-'Heightened conflict incidents'.
-'Severe job losses'.
-'Market instability'.
-'Investment crashes'.
-'Supply chain disruptions'.
-'Increased poverty rates'.
-'Currency devaluation'.
-'Overburdened local resources'.
-'Increased social tensions'.
-'Strained healthcare systems'.
-'Environmental degradation'.
-'Widespread famine outbreaks'.
-'Mass displacement waves'.
-'Severe medical shortages'.
-'Intensified poverty levels'.
\end{codex} 
  
\begin{codex} 
-'Severe biodiversity loss'.
-'Irreversible water pollution'.
-'Widespread unemployment surge'.
-'Investment capital evaporation'.
-'Consumer spending plummet'.
-'International trade paralysis'.
-'Financial sector instability'.
-'Intensified food shortages'.
-'Increased child mortality'.
\end{codex}

\EX
Note that the ``{\tt -}" operator marking negation is interpreted here as an unwanted outcome, from which, a rational agent would propagate back the denial of the initiator goal, implying that ``{\em escalation risks after use of tactical nuclear weapons}'' is something to avoid, given its consequences.

Similar initiator queries % with minor prompt tweaks
can cover recursive descent in consequences of things like unobservable / unmeasurable claims, undesirable outcomes of planned actions, implications of hidden legalese in contracts as well as untruthful advertisements or political persuasion hyperbolae.

By default, the DeepLLM system  \cite{flops24} generates propositional programs together with their unique minimal model\footnote{\url{https://github.com/ptarau/recursors}} and its low complexity polynomial solver can support scaling to very large programs aggregating positive and negative knowledge snippets consisting of facts and rules of Horn and Dual Horn programs. 
By asking an LLM to decompose recursively a task into subtasks organized as an AND-OR tree, the generated Prolog file will be a Horn clause program.
Dually, by asking an LLM to explore consequences of an undesirable state of the world or of a hypothesis that one would want to reject, the results will take the shape of a Dual Horn clause program.

This abundance of positive and negative information offers a fully explainable and semantically straightforward alternative to default reasoning and alleviates the contrast between the underlying open vs. closed world assumptions, given that arbitrary positive or negative information and constructive inference based on it is available on demand.

Explicit reasoning with negative information is also relevant for {\em unlearning} algorithms (in particular in-context unlearning) that ensure removal of stale information (e.g., who is the president of USA), untruthful content or copyrighted, toxic, dangerous, and otherwise harmful content (e.g.,  instructions on how to make napalm) \cite{liu2024unlearning}.

\section{Related Work}\label{rel}

Horn clause formulas and Dual Horn clause formulas (called weakly negative and, respectively, weakly positive in \cite{schaefer78}) are proven in his ``dichotomy theorem'' ( under the assumption $P \neq NP$) to be among the classes of propositional formulas that are in $P$. A graph-based linear algorithm exists for for Horn clause formulas, described in \cite{dowling} and given the linear renaming algorithm of propositional variables in \cite{renhorn90}, the same applies to Dual Horn formulas.
As a follow-up of \cite{dowling}, the Hornlog system, a logic programming alternative to Prolog covering (besides definite programs)  the handling of denial integrity constraints  has been implemented \cite{hornlog}. 

 A salient question
 %\footnote{Actually, asked back as part of a review request on a draft of this paper from Meta's Llama 3 LLM used as back-end by our {\em docdiver} system (see \url{https://docdiver.streamlit.app}).} 
one might ask is
``{\em If Dual Horn clauses have been known for a long time, why they haven't been widely used until now?}''.

A possible explanation is that the practical usefulness of negation as failure in Prolog and the related theoretical ramifications, culminating with the stable model semantics \cite{GelfondL88} and the emergence of Answer Set Programming as an alternative logic programming execution model have provided enough expressive power to deal with more subtle nuances of negative information.

Another is that constructive negation algorithms, going back to \cite{chan88,stuckey95} and  present in goal-directed ASP systems \cite{casp}, have provided, in the form of constraints on variable bindings, similar explanations covering the reasons of negative outcomes.
Also, a compilation scheme from normal logic programs to definite programs under given stratification constraints has been devised in \cite{stuckey90}.
Note however that constructive negation expressed as a set of ``is different from'' constraints on a set of variables is fundamentally weaker than inferring in Dual Horn clause logic what the actual values should be based on known negative facts.
 More precisely, in the case of classical constructive negation, these results were expressed as disjunction of conjunctions of the form \verb~X\=T~ where {\tt T} is a (usually ground) term, thus describing what {\tt X} cannot be, while in the case of Dual Horn programs the result is a positive binding of the form \verb~X=T~ saying what {\tt X} actually is. Thus, while the classical ``constructive'' negation in combination with constraint solvers provides an efficient filtering mechanism over a potentially infinite set of terms, it shares with negation as failure the fact that  it can only reject bindings constructed elsewhere in the program but it cannot actually {\em construct} variable bindings as its name would (somewhat inadvertently) suggest.

\section{Conclusion and Future Work}\label{conc}

We have revisited Dual Horn clauses as a ``back to the future'' endeavor,
motivated by their syntactic simplicity and straightforward semantics. As a result,
we have devised a compilation scheme that integrates them into Prolog programs. This enables our SymLP embedded language as a practical programming tool that reasons with explicitly specified positive and negative facts and rules. In  case of the possibly very large propositional programs generated by LLMs, the low polynomial complexity\footnote{actually linear if using the graph-based algorithm of \cite{dowling})} ensures tractability -- a key requirement for practical applications.

To some extent, the utility of Dual Horn program relies on the assumption that full knowledge of positive and negative facts can be acquired given their encapsulation in LLMs or traditional knowledge bases. This leaves open the possibility of unknown facts (e.g., those reached at the DeepLLM recursion depth limit) that can be seen as {\em abducibles}, i.e., verifiable if passing integrity constraints on the Horn clause side (formula (6)) or falsifiable if passing the corresponding constraints on the Dual Horn side (formula (8)).
Future work will be needed to study these in full detail, while aware that their presence, in the propositional case will keep  complexity polynomial.

This also opens the possibility to rely on soft-unification \cite{soft_unif}, relevant especially in the presence of LLM-generated facts stored as embeddings into a vector database. For instance, closeness via cosine-similarity to positive or negative facts could decide rules on which side would adopt these facts as abducibles, another future work direction worth to be explored.

\bibliographystyle{splncs}

\bibliography{tarau,ml,proglang, theory}

\end{document}

%% file: pheader24.tex
\usepackage{graphicx}
\usepackage{mathptmx}
\usepackage{tikz-qtree}
\usepackage{filecontents}
\usepackage{csvsimple}
\usepackage{amsfonts,amsmath}
\usepackage{subfig} % creates problems - fails maketitile
\usepackage{url}
\usepackage{verbatim}
\usepackage{color}
\usepackage{amsmath} 
\usepackage{proof}

\renewcommand{\phi}{\varphi}

\input prolog.tex

\definecolor{lgray}{gray}{0.95}
\definecolor{lblue}{rgb}{0.90,0.90,1.00}
\definecolor{lyellow}{rgb}{1.00,1.00,0.70}

\usepackage{listings}
\newtheorem{prop}{Proposition}
\newtheorem{ex}{Example}

\lstnewenvironment{codex}
    {\lstset{}%
      \csname lst@SetFirstLabel\endcsname}
    {\csname lst@SaveFirstLabel\endcsname}
\newcommand{\BI}[0]{\begin{itemize}}
\newcommand{\EI}[0]{\end{itemize}}
\newcommand{\I}[0]{\item}
\newcommand{\BE}[0]{\begin{enumerate}}
\newcommand{\EE}[0]{\end{enumerate}}

\newcommand{\BX}[0]{\begin{ex}}
\newcommand{\EX}[0]{\end{ex}}
\newcommand{\BV}[0]{\begin{verbatim}}
\newcommand{\EV}[0]{\end{verbatim}}

\newcommand{\BP}[0]{\begin{prop}}
\newcommand{\EP}[0]{\end{prop}}

\newcommand{\BEQ}{\begin{equation}}
\newcommand{\EEQ}{\end{equation}}

\newcommand{\IF}{\leftarrow}
\newcommand{\FI}{\rightarrow}
\newcommand{\AND}{\land}
\newcommand{\OR}{\vee}
\newcommand{\NOT}{\neg}

\newcommand{\BC}[0]{\begin{center}}
\newcommand{\EC}[0]{\end{center}}

\newcommand{\BF}[0]{\begin{filecontents*}{data.csv}}

\newcommand{\BQ}[0]{\color{blue}\begin{quote}}
\newcommand{\EQ}[0]{\end{quote}\color{black}}

\def \bscale1 {0.25}
\def \bscale {0.25}

%% file: dual_arxiv.bbl
\begin{thebibliography}{10}
\providecommand{\url}[1]{\texttt{#1}}
\providecommand{\urlprefix}{URL }
\providecommand{\doi}[1]{https://doi.org/#1}

\bibitem{casp}
ARIAS, J., CARRO, M., SALAZAR, E., MARPLE, K., GUPTA, G.: Constraint answer set
  programming without grounding. Theory and Practice of Logic Programming
  \textbf{18}(3-4),  337--354 (2018). \doi{10.1017/S1471068418000285}

\bibitem{chan88}
Chan, D.: {Constructive Negation Based on the Completed Database}. In:
  Kowalski, R.A., Bowen, K.A. (eds.) Logic Programming, Proceedings of the
  Fifth International Conference and Symposium, Seattle, Washington, USA,
  August 15-19, 1988 {(2} Volumes). pp. 111--125. {MIT} Press (1988)

\bibitem{renhorn90}
Chandru, V., Coullard, C.R., Hammer, P.L., Monta\~{n}ez, M., Sun, X.: {On
  renamable Horn and generalized Horn functions}. Annals of Mathematics and
  Artificial Intelligence  \textbf{1}(1-4),  33--47 (sep 1990).
  \doi{10.1007/BF01531069}, \url{https://doi.org/10.1007/BF01531069}

\bibitem{soft_unif}
Cingillioglu, N., Russo, A.: Learning invariants through soft unification. In:
  Larochelle, H., Ranzato, M., Hadsell, R., Balcan, M., Lin, H. (eds.) Advances
  in Neural Information Processing Systems 33: Annual Conference on Neural
  Information Processing Systems 2020, NeurIPS 2020, December 6-12, 2020,
  virtual (2020),
  \url{https://proceedings.neurips.cc/paper/2020/hash/5d0d5594d24f0f955548f0fc0ff83d10-Abstract.html}

\bibitem{dantsin}
Dantsin, E., Eiter, T., Gottlob, G., Voronkov, A.: Complexity and expressive
  power of logic programming. ACM Comput. Surv.  \textbf{33}(3),  374–425
  (sep 2001). \doi{10.1145/502807.502810},
  \url{https://doi.org/10.1145/502807.502810}

\bibitem{dowling}
Dowling, W.F., Gallier, J.H.: {Linear-Time Algorithms for Testing the
  Satisfiability of Propositional Horn Formulae.} J. Log. Program.
  \textbf{1}(3),  267--284 (1984),
  \url{https://doi.org/10.1016/0743-1066(84)90014-1}

\bibitem{dy1}
Dyckhoff, R.: {Contraction-free sequent calculi for intuitionistic logic}.
  Journal of Symbolic Logic  \textbf{57}(3),  795--807 (1992).
  \doi{10.2307/2275431}

\bibitem{eiter2007complexity}
Eiter, T., Fink, M., Tompits, H., Woltran, S.: {Complexity Results for Checking
  Equivalence of Stratified Logic Programs.} In: IJCAI. pp. 330--335 (2007)

\bibitem{eshghi89}
Eshghi, K., Kowalski, R.A.: {Abduction Compared with Negation by Failure}. In:
  Levi, G., Martelli, M. (eds.) Logic Programming, Proceedings of the Sixth
  International Conference, Lisbon, Portugal, June 19-23, 1989. pp. 234--254.
  {MIT} Press (1989)

\bibitem{hornlog}
Gallier, J.H., Raatz, S.: {HORNLOG: A graph-based interpreter for general Horn
  clauses}. The Journal of Logic Programming  \textbf{4}(2),  119--155 (1987).
  \doi{https://doi.org/10.1016/0743-1066(87)90015-X},
  \url{https://www.sciencedirect.com/science/article/pii/074310668790015X}

\bibitem{GelfondL88}
Gelfond, M., Lifschitz, V.: The stable model semantics for logic programming.
  In: Kowalski, R.A., Bowen, K.A. (eds.) Logic Programming, Proceedings of the
  Fifth International Conference and Symposium, Seattle, Washington, USA,
  August 15-19, 1988 {(2} Volumes). pp. 1070--1080. {MIT} Press (1988)

\bibitem{stuckey90}
Kanchanasut, K., Stuckey, P.: Eliminating negation from normal logic programs.
  In: Kirchner, H., Wechler, W. (eds.) Algebraic and Logic Programming. pp.
  217--231. Springer Berlin Heidelberg, Berlin, Heidelberg (1990)

\bibitem{liu2024unlearning}
Liu, K.Z.: Machine unlearning in 2024 (Apr 2024),
  https://ai.stanford.edu/~kzliu/blog/unlearning

\bibitem{miller21}
Miller, D.: A survey of the proof-theoretic foundations of logic programming.
  CoRR  \textbf{abs/2109.01483} (2021), \url{https://arxiv.org/abs/2109.01483}

\bibitem{popper34}
Popper, K.R.: The Logic of Scientific Discovery. Hutchinson, London (1934)

\bibitem{schaefer78}
Schaefer, T.J.: The complexity of satisfiability problems. In: Proceedings of
  the Tenth Annual ACM Symposium on Theory of Computing. p. 216?226. STOC '78,
  Association for Computing Machinery, New York, NY, USA (1978).
  \doi{10.1145/800133.804350}, \url{https://doi.org/10.1145/800133.804350}

\bibitem{stuckey95}
Stuckey, P.: {Negation and Constraint Logic Programming}. Information and
  Computation  \textbf{118}(1),  12--33 (1995).
  \doi{https://doi.org/10.1006/inco.1995.1048},
  \url{https://www.sciencedirect.com/science/article/pii/S0890540185710486}

\bibitem{tplp22}
Tarau, P.: {Abductive Reasoning in Intuitionistic Propositional Logic via
  Theorem Synthesis}. Theory and Practice of Logic Programming  \textbf{22}(5),
   693--707 (2022). \doi{10.1017/S1471068422000254}

\bibitem{tarau2023automation}
{Tarau}, P.: {Full Automation of Goal-driven LLM Dialog Threads with And-Or
  Recursors and Refiner Oracles} arXiv:2306.14077 (Jun 2023).
  \doi{10.48550/arXiv.2306.14077}

\bibitem{flops24}
{Tarau}, P.: {System Description: DeepLLM, Casting Dialog Threads into Logic
  Programs}. In: Gibbons, J., Miller, D. (eds.) {Proceedings of 17th
  International Symposium on Functional and Logic Programming (FLOPS 2024)}.
  Springer LNCS 14659 (2024)

\end{thebibliography}
